\documentclass[pdflatex,sn-basic]{sn-jnl}

\jyear{2024}

\theoremstyle{thmstyleone}

\theoremstyle{thmstyletwo}

\theoremstyle{thmstylethree}

\usepackage{tabularx}
\newcolumntype{L}[1]{>{\raggedright\let\newline\\\arraybackslash\hspace{0pt}}p{#1}}
\newcolumntype{C}[1]{>{\centering\let\newline\\\arraybackslash\hspace{0pt}}p{#1}}
\newcolumntype{R}[1]{>{\raggedleft\let\newline\\\arraybackslash\hspace{0pt}}b{#1}}

\begin{document}

\title[KD \& IG]{Knowledge Distillation: Enhancing Neural Network Compression with Integrated Gradients}

\author[1]{\fnm{David E.} \sur{Hernandez}, \dgr{B.S.}}\email{david.hernandez@nordlinglab.org (0009-0004-4392-6927)}

\author[1]{\pfx{Dr} \fnm{Jose Ramon} \sur{Chang}, \dgr{MSc, PhD}}\email{jose.chang@nordlinglab.org (0000-0001-5587-7828)}

\author*[1]{\pfx{Dr} \fnm{Torbj{\"o}rn E. M.} \sur{Nordling}, \dgr{MSc, PhD}}\email{torbj{\"o}rn.nordling@nordlinglab.org (0000-0003-4867-6707)}

\affil[1]{\orgdiv{Department of Mechanical Engineering}, \orgname{National Cheng Kung University}, \orgaddress{\street{No. 1 University Rd.}, \city{Tainan} \postcode{701}, \country{Taiwan}}}

\abstract{
Efficient deployment of deep neural networks on resource-constrained devices demands advanced compression techniques that preserve accuracy and interpretability.
This paper proposes a machine learning framework that augments Knowledge Distillation (KD) with Integrated Gradients (IG), an attribution method, to optimise the compression of convolutional neural networks.
We introduce a novel data augmentation strategy where IG maps, precomputed from a teacher model, are overlaid onto training images to guide a compact student model toward critical feature representations.
This approach leverages the teacher’s decision-making insights, enhancing the student’s ability to replicate complex patterns with reduced parameters.
Experiments on CIFAR-10 demonstrate the efficacy of our method: a student model, compressed 4.1-fold from the MobileNet-V2 teacher, achieves 92.5\% classification accuracy, surpassing the baseline student’s 91.4\% and traditional KD approaches, while reducing inference latency from 140 ms to 13 ms--a tenfold speedup.
We perform hyperparameter optimisation for efficient learning.
Comprehensive ablation studies dissect the contributions of KD and IG, revealing synergistic effects that boost both performance and model explainability.
Our method’s emphasis on feature-level guidance via IG distinguishes it from conventional KD, offering a data-driven solution for mining transferable knowledge in neural architectures.
This work contributes to machine learning by providing a scalable, interpretable compression technique, ideal for edge computing applications where efficiency and transparency are paramount.
}


\maketitle

\section{Introduction}\label{sec:introduction}\label{sec:intro}

The implementation of Deep Learning (DL) architectures within constrained computing environments presents significant technical challenges, particularly for edge computing platforms like mobile devices and embedded systems \citep{Szegedy2017Inceptionv4IA, deng2020model, krishnamoorthi2018quantizing}. 
Contemporary neural network architectures frequently exceed the operational capabilities of these platforms, resulting in performance degradation through increased computational latency, elevated power consumption, expanded memory utilisation, and potential data security vulnerabilities \citep{han2015deep, howard2017mobilenets, sze2017efficient}.

Advanced foundation models such as GPT-4 and DeepSeek-R1 demonstrate exceptional cognitive capabilities but impose substantial computational requirements, emphasising the necessity for optimised model compression methodologies in resource-constrained operational contexts \citep{chen2024smallmodels}.
This situation illuminates a significant research opportunity: while considerable scholarly effort has been directed toward advancing large-scale architectures, systematic investigation into compression technique refinement for practical implementation remains insufficiently explored.

Model compression techniques can enable efficient deployment without severe performance loss \citep{cheng2017survey}.
Among various approaches, Knowledge Distillation (KD) offers particular advantages by transferring knowledge from a large teacher model to a smaller student model using soft targets \citep{Hinton2015}. 
Meanwhile, Integrated Gradients (IG), an explainable AI technique, attributes feature importance at the pixel level, enhancing interpretability \citep{sundararajan2017axiomatic}.

CIFAR-10 benchmark evaluations have demonstrated the effectiveness of various distillation approaches. 
Studies by \citep{zhao2020highlight} and \citep{choi2020data} have explored adaptative distillation strategies that dynamically adjust knowledge transfer based on image complexity, achieving improved performance on challenging samples within the CIFAR-10 dataset. 
Meanwhile, \citep{bhardwaj2019memory} demonstrated that memory-efficient distillation can achieve up to 19x compression while maintaining over 94\% accuracy on CIFAR-10. 
These approaches, however, typically do not leverage model interpretability mechanisms to enhance the compression process.

Our research addresses this opportunity through the development of an augmented knowledge distillation framework that integrates IG with KD to improve compression. 
By guiding the student with pre-computed IG maps as data augmentation, this approach aims to preserve model interpretability while achieving substantial parameter reduction. 
This method targets edge computing applications, balancing efficiency, accuracy, and explainability.

The structural organisation of this manuscript proceeds as follows: Section \ref{sec:methodology} provides a comprehensive examination of our methodological approach, encompassing the teacher-student architectural paradigm, knowledge distillation principles, and the strategic integration of integrated gradients for data augmentation purposes. It delineates our experimental protocol, including dataset selection criteria and parameter optimisation strategies for knowledge distillation.
Section \ref{sec:results} presents quantitative and qualitative experimental findings, demonstrating the effectiveness of our proposed techniques using MobileNet-V2 architecture evaluated on the CIFAR-10 benchmark dataset.
Finally, Section \ref{sec:conclusions} summarises key contributions and examines potential research trajectories in neural network compression for edge computing applications, alongside broader implications of our findings for the field.

\subsection{Explainable AI} \label{sec:XAI}
The field of explainable artificial intelligence (XAI) encompasses methodologies that enhance transparency in machine learning systems, making their decision processes more accessible to human understanding.
Among the diverse approaches developed for model interpretation, including contrastive explanations and surrogate modelling techniques, our research specifically focuses on feature attribution through integrated gradients visualisation, as this method offers particular advantages for compression guidance.

Attribution techniques quantify the contribution of individual input elements to specific model outputs. 
The IG method, developed as a principled approach to feature importance calculation, accumulates gradient information along a parameterised path from a reference point to the input sample. 
This approach satisfies key theoretical properties including sensitivity and implementation invariance, while producing intuitively meaningful visualisations of feature relevance \citep{sundararajan2017axiomatic}.

Transparency becomes particularly significant in compressed models intended for critical applications, where understanding decision rationales remains essential despite architectural simplification. 
In healthcare applications, for instance, clinicians require confidence that simplified models operating on mobile devices maintain the same diagnostic focus as their comprehensive counterparts, e.g. identify the same region of tumour cells in a biopsy slice.

Our research introduces a novel integration of attribution mechanisms with knowledge distillation frameworks. 
By incorporating IG as a data augmentation technique, we provide the student model with explicit feature-level guidance during training. 
This approach addresses limitations in conventional compression methodologies by directly leveraging interpretability to enhance the optimisation process, improving both functional performance and the transparency of model decisions.

\subsection{Previous Works} \label{sec:lit_review}
Model compression techniques like pruning and quantisation reduce network size but often compromise accuracy at high compression rates \citep{cheng2017survey}.
KD mitigates this issue by training a student model on the softened outputs of the teacher, preserving decision-making capabilities \citep{Hinton2015}.
Studies show the effectiveness of KD in maintaining performance across tasks \citep{chen2024smallmodels}.

Our literature study focuses on model compression through knowledge distillation in image classification on CIFAR-10.
The studies revealed a complex relationship between compression factor and accuracy, with reported compression in the range (1.9x to 20.8x) and accuracy impacts in the range (-8.63\% to +0.30\%).

\begin{figure}[t]
  \centering
  \includegraphics[width=\linewidth]{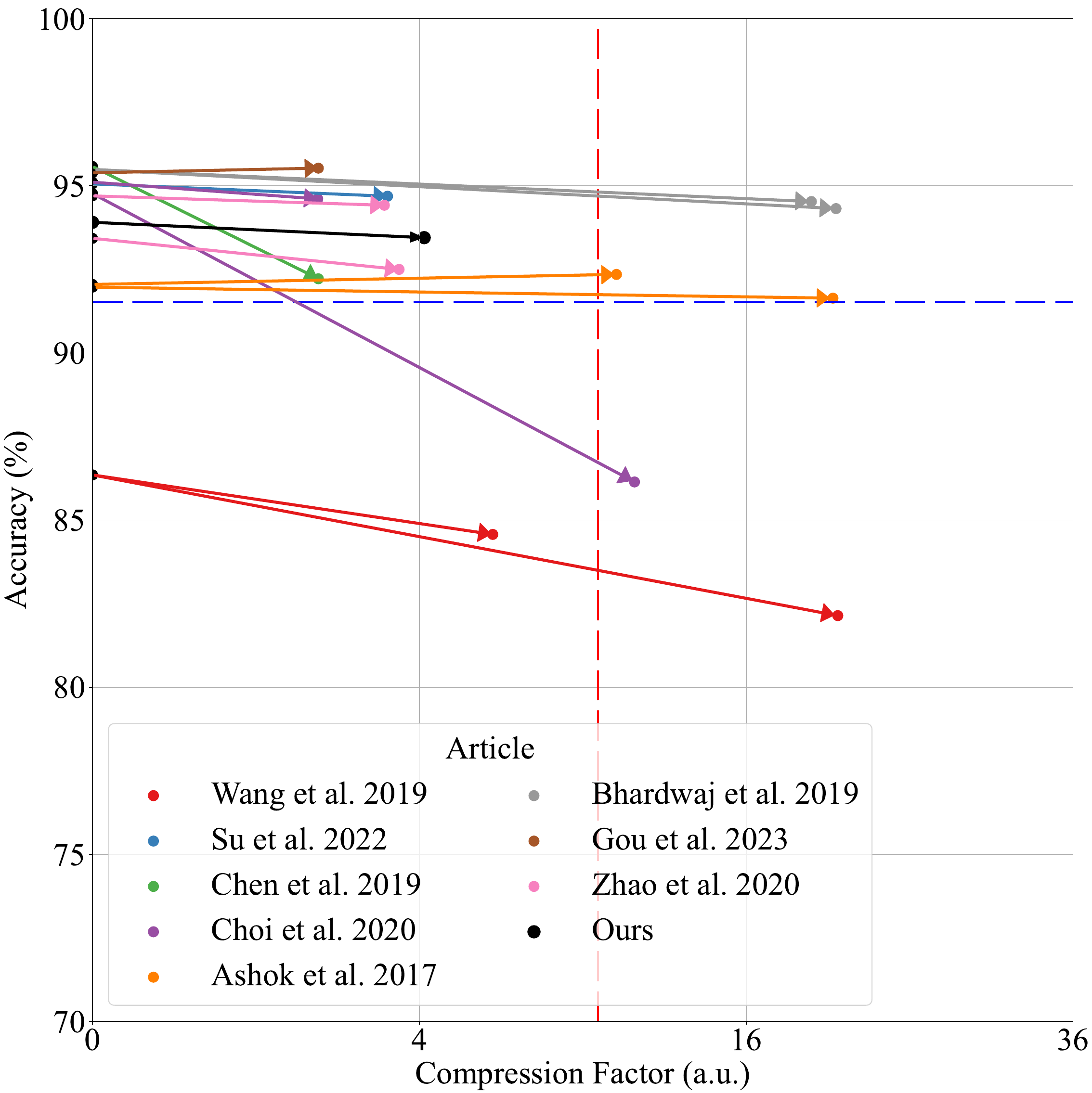}
  \caption{Relationship between compression factor and accuracy across different knowledge distillation approaches on CIFAR-10. 
  Each line connects a teacher model (left point) to its compressed student counterpart (right point). 
  Our IG-enhanced KD approach (in black) achieves competitive accuracy at 4.1x compression, balancing efficiency and performance. 
  The blue dashed line represents the mean accuracy (91.51\%) across studies, while the red dashed line indicates the mean compression factor (9.57x).}
  \label{fig:plot_studies}
\end{figure}

Table~\ref{tab:paper_results} and Fig.~\ref{fig:plot_studies} reveals the relationship between compression factors and accuracy changes across different studies, including our proposed approach. 
Trade-offs between model size reduction and performance are visible, with studies like \cite{choi2020data} showing significant variation in outcomes even within their own experiments. 
While they achieved minimal accuracy loss (-0.50\%) at 1.9x compression, pushing to 11x compression resulted in more substantial degradation (-8.63\%).

In contrast, other researchers have demonstrated remarkable efficiency gains with minimal performance impact. 
\cite{bhardwaj2019memory} achieved 19.35x compression while maintaining 94.53\% accuracy, just 0.96 percentage points below their teacher model. 
\cite{gou2023hierarchical} demonstrated that well-designed knowledge transfer could even improve performance in some cases, with their student model exceeding teacher accuracy by 0.14 percentage points despite 1.91x compression.
This, in our opinion, indicate a potential to improve the teacher and could simply be due to suboptimal training of the teacher.

The architectural approaches to compression vary considerably across studies, as shown in Table~\ref{tab:paper_results2}.
Some researchers adopted standardised model pairs like ResNet-34 to ResNet-18 \citep{chen2019data, gou2023hierarchical}, while others explored custom architectures or systematic parameter reduction through layer removal \citep{ashok2017n2n}.
The Wide ResNet (WRN) architecture family features prominently, with several studies compressing WRN-40 variants to their WRN-16 counterparts \citep{su2022stkd, choi2020data, zhao2020highlight}.

\begin{table}[bt]
\centering
\caption{Results comparing teacher and student model accuracies across CIFAR10. 
Only the highest accuracy and highest compression results for each study were included. 
All accuracies are assumed to be testing accuracies unless noted with an asterisk (*), which indicates training accuracies, ``Baseline" refers to the accuracy of the student model before any knowledge distillation has been applied, while ``Distilled" denotes the accuracy after applying the distillation process. 
``Rel. $\Delta$Acc. means the Relative $\Delta$Accuracy between the Teacher, Student baseline, and Student distilled accuracies. The corresponding model parameters are detailed in Table~\ref{tab:paper_results2}.}
\label{tab:paper_results}
\begin{tabular}{p{1.3cm}p{1cm}p{1cm}p{1cm}p{1cm}p{1.5cm}p{2.4cm}}
\hline
Teacher  & \multirow{2}{*}{CF} & \multicolumn{2}{c}{Student Acc. (\%)} & \multirow{2}{*}{$\Delta$Acc.} & Rel.  & \multirow{2}{*}{Article}                                \\
Acc. (\%)               &                     & Baseline          & Distilled         &                               & $\Delta$Acc. (\%) &                                                         \\ \hline
\multirow{2}{*}{86.35*} & 6.00                & -                 & 84.57*            & -1.78                         & -                 & \multirow{2}{*}{\citet{wang2019private}}   \\
                        & 20.80               & -                 & 82.14*            & 4.21                          & -                 &                                                         \\
95.06                   & 3.26                & 93.50             & 94.69             & -0.37                         & 76.28             & \citet{su2022stkd}                         \\
95.58                   & 1.91                & 93.92             & 92.22             & -3.36                         & -102.41           & \citet{chen2019data}                    \\
95.11                   & 1.90                & 94.56             & 94.61             & -0.50                         & 9.09              & \multirow{2}{*}{\citet{choi2020data}}      \\
94.77                   & 11.00               & 90.97             & 86.14             & -8.63                         & -127.10           &                                                         \\
92.05                   & 10.28               & -                 & 92.35             & +0.30                         & -                 & \multirow{2}{*}{\citet{ashok2017n2n}}      \\
91.97                   & 20.53               & -                 & 91.64             & -0.33                         & -                 &                                                         \\
\multirow{2}{*}{95.49}  & 19.35               & -                 & 94.53             & -0.96                         & -                 & \citet{bhardwaj2019memory}             \\
                        & 20.7                & -                 & 94.32             & -1.17                         & -                 &                                                         \\
87.32                   & 0.00                & -                 & 86.59             & -0.73                         & -                 & \citet{blakeney2020parallel}             \\
95.39                   & 1.91                & 94.93             & 95.53             & +0.14                         & 130.43            & \citet{gou2023hierarchical}                \\
95.01                   & -                   & 91.03             & 93.58             & -1.43                         & 64.07             &                                                         \\
94.70                   & 3.19                & 93.68             & 94.42             & -0.30                         & 72.55             & \multirow{2}{*}{\citet{zhao2020highlight}} \\
93.43                   & 3.52                & 91.28             & 92.50             & -0.93                         & 56.74             &                  \\                                      
\hline
\end{tabular}
\end{table}

\begin{table}[tb]
\centering
\caption{Parameters of the teacher and student models corresponding to the accuracies reported in Table~\ref{tab:paper_results}.  }
\label{tab:paper_results2}
\begin{tabular}{p{1.6cm}p{1.5cm}p{2cm}p{1.5cm}p{2.8cm}}
\hline
Teacher Model             & $\#$ Params            & Student Model              & $\#$ Params             & Article                                                   \\ \hline
-                         & \multirow{2}{*}{3.12M} & -                          & 0.52M                   & \multirow{2}{*}{\citet{wang2019private}}     \\
-                         &                        & -                          & 0.15M                   &                                                           \\
WRN-40-2                  & 2.25M                  & WRN-16-2                   & 0.69M                   & \citet{su2022stkd}                          \\
ResNet-34                 & $\sim$21M              & ResNet-18                  & $\sim$11M               & \citet{chen2019data}                         \\
ResNet-34                 & 21.3M                  & ResNet18                   & 11.2M                   & \multirow{2}{*}{\citet{choi2020data}}        \\
WRN-40-2                  & 2.2M                   & WRN-16-1                   & 0.2M                    &                                                           \\
ResNet-34                 & 21.28M                 & Layer removal \& shrinkage & 2.07M      & \multirow{4}{*}{\citet{ashok2017n2n}}        \\
VGG19                     & 20.2M                  & Layer removal \& shrinkage & 984k &                   \\
\multirow{2}{*}{WRN-40-4} & \multirow{2}{*}{8.9M}  & NoNN-2S-XL                 & 0.46M                   & \multirow{2}{*}{\citet{bhardwaj2019memory}}  \\
                          &                        & NoNN-2S                    & 0.43M                   &                                                           \\
VGG16                     & 138.36M                & VGG16                      & 138.36M                 & \citet{blakeney2020parallel}                 \\
ResNet-34                 & 21.30M                 & ResNet-18                  & \multirow{2}{*}{11.17M} & \multirow{2}{*}{\citet{gou2023hierarchical}} \\
ResNet-18                 & 11.17M                 & ShuffleNet-v2              &                         &                                                           \\
WRN-40-2                  & 2.20M                  & WRN-16-2                   & 0.69M                   & \multirow{2}{*}{\citet{zhao2020highlight}}   \\
WRN-40-1                  & 0.57M                  & WRN-16-1                   & 0.17M                   &  \\
\hline
\end{tabular}
\end{table}

Recent advancements have emphasised refined distillation methodologies to manage these competing considerations.
\cite{gou2023hierarchical} proposed a multi-level distillation framework, providing enhanced precision in balancing compression against accuracy.
\cite{choi2020data} investigated flexible distillation approaches that automatically calibrate according to the task's difficulty, demonstrating diverse outcomes when tested on CIFAR-10.
\cite{zhao2020highlight} focused on feature highlighting, guiding student models to mimic teacher attention on important regions, achieving 94.42\% accuracy at 3.19x compression.

IG, introduced by \citet{sundararajan2017axiomatic}, provides fine-grained feature attributions and has been widely used for interpretability in image classification \citep{ribeiro2016lime}. 
However, only \cite{wu2023ad} have in the loss function used IG to transfer knowledge in an NLP task.
Unlike pruning or quantisation, which directly modify network structure, our approach uses IG to enhance KD by focusing on feature guidance. 
This differs from traditional KD by emphasising data-driven knowledge transfer, avoiding direct parameter reduction trade-offs.

Our research builds upon these approaches but introduces a fundamentally different mechanism for knowledge transfer. 
While previous methods have primarily focused on output distributions or intermediate feature representations, our approach leverages IG to provide explicit guidance about feature importance at the pixel level. 
As indicated in Fig.~\ref{fig:plot_studies}, our method achieves competitive accuracy at a moderate compression factor, balancing performance and efficiency. 
The specific details of our approach and comprehensive results will be presented in subsequent sections.

\vspace{.3cm}
This paper makes the following contributions:

\begin{itemize}
    \item Proposed a novel model compression method that uniquely integrates IG with KD, using attribution maps to guide feature learning in student networks.
    
    \item Introduced an efficient implementation strategy that pre-computes IG attribution maps before training begins, transforming what would be a prohibitive computational cost during training into a one-time preprocessing step.
    
    \item Provided systematic ablation studies that isolate the individual and combined effects of KD and IG, revealing their complementary benefits for model compression.
\end{itemize}

\section{Methodology} \label{sec:methodology}
\begin{figure}[tb]
  \centering
  \includegraphics[width=\linewidth]{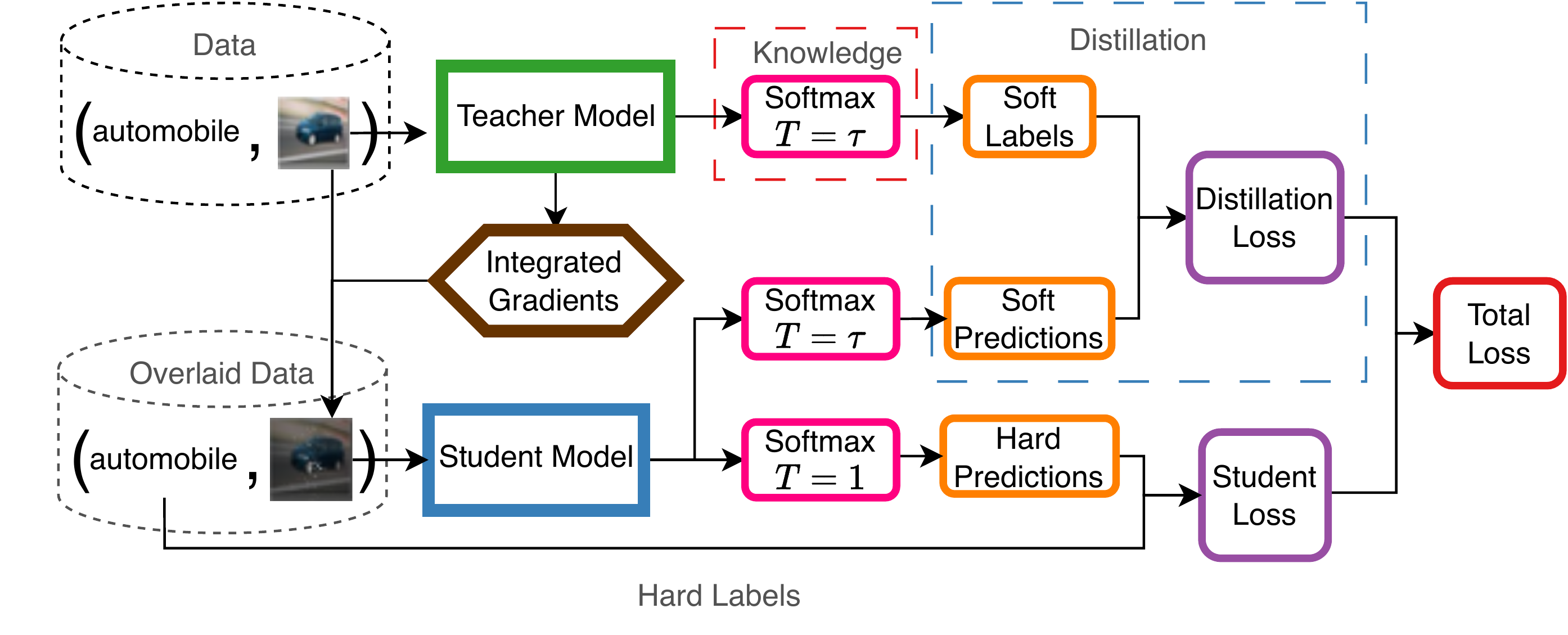}
  \caption{Knowledge distillation process using integrated gradients for data augmentation. 
  The teacher model (green) employs a temperature parameter $T=\tau$ where $\tau > 1$ in its softmax function to produce soft targets, which, along with the hard labels from the dataset, guide the training of the student model (blue). 
  Integrated gradients (brown) are overlaid with the original images to generate enhanced data that focuses critical features that the student model should use during training.
}
  \label{fig:kd_ig_flow}
\end{figure}

Our proposed compression framework integrates feature attribution with knowledge transfer to create efficient yet high-performing neural networks. 
The approach leverages two complementary components: knowledge distillation and integrated gradients-based data augmentation. The process is visualised in Fig.~\ref{fig:kd_ig_flow}

\subsection{Compression Through Knowledge Distillation}
The core of our compression strategy relies on distilling knowledge from a complex teacher network into a compact student network. 
Unlike standard supervised learning that relies solely on hard target labels, knowledge distillation captures the nuanced decision boundaries of the teacher through probability distributions.

During training, the teacher model--MobileNetV2, 2.2M parameters, \citep{howard2017mobilenets}--processes each input to generate soft targets with temperature parameter $T$. 
These soft targets preserve crucial information about similarity relationships between classes that hard labels discard. 
For example, when classifying an image of a dog, the teacher might assign meaningful probabilities to visually similar classes like wolves or foxes.

Our objective function balances two learning signals:
\begin{align} \label{eq:weighted_loss}
L_{KD} = (1 - \alpha)L_\mathcal{H} + \alpha L_\mathcal{KL}.
\end{align}

\noindent Here $L_\mathcal{H}$ represents standard cross-entropy loss with ground truth (hard labels):
\begin{align}
L_\mathcal{H} = -\sum_{i}y_i\log(f_s(x)_i)
\end{align}
and $L_\mathcal{KL}$ measures Kullback-Leibler divergence between teacher and student output distributions, (i.e., the soft labels and soft predictions):
\begin{align}
L_\mathcal{KL} = \sum_{i}f_t(x; T)_i\log\left(\frac{f_t(x; T)_i}{f_s(x; T)_i}\right).
\end{align}

\noindent Here, $y$ represents one-hot encoded labels, $f_t$ and $f_s$ are teacher and student models respectively for the input image $x$, and $T$ is the temperature parameter controlling the softness of probability distributions. 
The hyperparameter $\alpha$ controls the relative importance of these signals, with optimal values determined through systematic experimentation.

\subsection{Feature-Guided Learning with Integrated Gradients}
While knowledge distillation transfers output-level knowledge, it provides limited guidance about which input features influenced the decisions of the teacher. 
To address this limitation, we introduce a novel data augmentation strategy using integrated gradients (IG).

IG quantifies the contribution of each input pixel to a specific prediction by computing:
\begin{align}
IG_i(x) = (x_i - x'_i) \int_{\beta=0}^{1} \frac{\partial F(x' + \beta(x-x'))}{\partial x_i}d\beta.
\end{align}

\noindent Here $x$ represents the input image, $x'$ a baseline input (typically black), and $F$ the prediction function of the teacher. 
This gradient path integral satisfies desirable theoretical properties including sensitivity and implementation invariance.

Our innovation lies in how these attribution maps guide the learning of the student. 
Rather than incorporating IG into the loss function, we overlay IG maps onto training images with probability $p$, creating augmented inputs that visually emphasise features the teacher deemed important. 

We implement IG as data augmentation, overlaying IG maps onto images with probability $p$:
\begin{align}
IG_{\text{scaled}}(x) &= IG(x)^s \quad \text{with } s \sim \exp(\mathcal{U}[\ln(1), \ln(2)]), \\
\hat{IG}(x) &= \frac{IG_{\text{scaled}}(x) - \min(IG_{\text{scaled}}(x))}{\max(IG_{\text{scaled}}(x)) - \min(IG_{\text{scaled}}(x))}, \\
x_{\text{augmented}} &= 
\begin{cases} 
0.5 \cdot x + 0.5 \cdot \hat{IG} & \text{with probability } p, \\ 
x & \text{otherwise}.
\end{cases}
\end{align}
The scale factor $s$ is drawn from a log-uniform distribution, and the overlay blends $x$ and $\hat{IG}$ equally when applied, putting emphasis on high-IG regions.

To optimise this process, we (1) pre-compute IG maps for all training images before student training begins, converting a potential computational bottleneck into a one-time preprocessing step, (2) apply logarithmic scaling to control attribution intensity, focusing on the most influential regions, (3) implement stochastic overlay with probability $p$ to prevent overfitting to attribution patterns.

This data-driven approach to knowledge transfer complements traditional distillation by providing pixel-level guidance that helps the student efficiently learn from limited parameters. Crucially, this guidance preserves model interpretability by ensuring the student focuses on the same discriminative features as the teacher.

\subsection{Experimental Setup}
We conducted our experiments on the CIFAR-10 dataset, comprising 50,000 training and 10,000 test $32 \times 32$ colour images across 10 classes. 
Our teacher model is MobileNet-V2 with 2.2M parameters, achieving 93.91\% baseline accuracy on the test set.

The student model was systematically compressed by reducing both depth and width, resulting in a 4.1x reduction to 543,498 parameters. 
This architecture preserves early feature extraction layers while removing deeper layers, maintaining output feature dimensions sufficient for classification.

Training was performed for 100 epochs using the Adam optimiser with a learning rate of 0.001.

All experiments were conducted on an NVIDIA RTX 3090 GPU, with performance metrics including accuracy and inference time to quantify functional performance.

\section{Results and Discussion}\label{sec:results}
Our experiments evaluated the effectiveness of both individual and combined approaches to model compression, focusing specifically on KD and IG augmentation.

\subsection{Performance Comparison}
Table~\ref{tab:accuracy} presents the test accuracy comparison across different model configurations. 
The teacher model (MobileNetV2) achieved 93.91\% accuracy on CIFAR-10, while the baseline student model with 4.1x fewer parameters reached 91.43\%, representing a 2.48 percentage point degradation.

\begin{table}[t!]
\centering
\caption{Test accuracy comparison across different methods for the student model with 4.1x compression.}
\label{tab:accuracy}
\begin{tabular}{lcc}
\hline
Method & Accuracy (\%) & $\Delta$Acc. vs Baseline \\
\hline
Teacher & 93.91 & - \\
Baseline Student & 91.43 & - \\
KD & 92.29 & +0.86 \\
IG & 92.01 & +0.58 \\
KD \& IG & \textbf{92.45} & \textbf{+1.02} \\
\hline
\end{tabular}
\end{table}

Applying knowledge distillation significantly improved performance, with the KD-trained student achieving 92.29\% accuracy, reducing the gap from the teacher by approximately one-third. 
Remarkably, our IG-based augmentation approach independently improved the accuracy of the student model to 92.01\%, demonstrating the value of feature attribution guidance even without traditional distillation.

The most significant finding is that combining KD with IG yields superior results, reaching 92.45\% accuracy—a 1.02 percentage point improvement over the baseline student. 
This combined approach preserves 98.4\% of the performance of the teacher model while using only 24.3\% of the parameters.

\subsection{Computational Efficiency}
Beyond accuracy preservation, our compressed model demonstrates substantial gains in computational efficiency. Table~\ref{tab:latency} shows that the 4.1x parameter reduction translates to a 10.8x reduction in inference time, from 140 ms to 13 ms per batch of images on our test hardware.

\begin{table}[t!]
\centering
\caption{Inference latency comparison between teacher and student models.}
\label{tab:latency}
\begin{tabular}{lcc}
\hline
Model & Latency (ms) & Speedup \\
\hline
Teacher & 140 & 1.0x \\
Student & 13 & 10.8x \\
\hline
\end{tabular}
\end{table}

This non-linear improvement in computational efficiency likely results from reduced memory access patterns and better cache utilisation in the smaller model, highlighting the practical advantages of model compression for real-time applications.

\subsection{Optimal Hyperparameters}
We did a systematic grid search to identify optimal hyperparameters for both knowledge distillation and integrated gradients, evaluating 5-9 values of $T$, $\alpha$, and $p$.  
For KD, a moderate temperature of $T=2.5$ achieved the best balance between preserving class relationships and maintaining categorical distinction, while a relatively low distillation weight of $\alpha=0.01$ provided effective guidance without overwhelming the learning from hard labels.

For the IG overlay, a probability of $p=0.1$ emerged as optimal, balancing feature emphasis with model generalisation. 
Higher probabilities (0.25, 0.5) led to performance degradation, suggesting excessive attribution information may cause overemphasis on specific features at the expense of learning diverse representations. All presented results are for these optimal parameters.

\subsection{Visualisation Analysis}
\begin{figure}
  \centering
  \includegraphics[width=\linewidth]{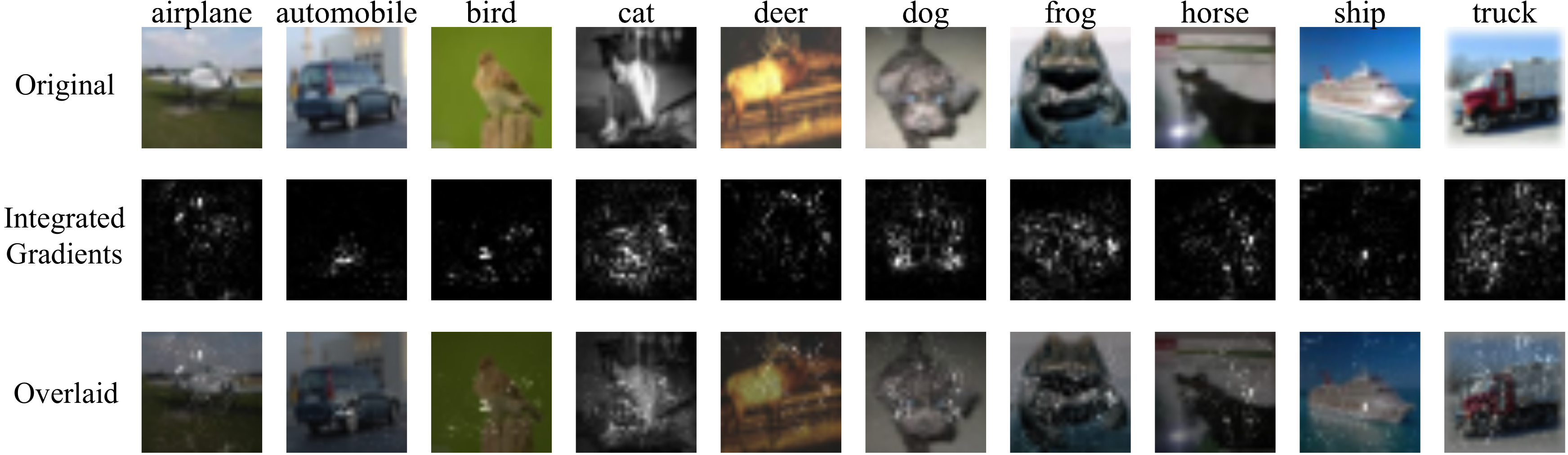}
  \caption{Implementation of IG as a data augmentation technique on CIFAR-10. 
  The top row shows original images from various classes. 
  The middle row displays the Integrated Gradients, highlighting areas significantly influencing the predictions of the teacher model. 
  The bottom row presents overlaid images, combining originals with their respective integrated gradients to emphasise regions of interest.
}
  \label{fig:augment_class}
\end{figure}

Fig.~\ref{fig:augment_class} illustrates our IG augmentation technique applied to CIFAR-10 samples. 
The visualisation reveals how IG maps highlight class-discriminative features, which are then overlaid onto original images to guide the attention of the student model. 
For example, in the automobile image, IG emphasises the wheels and body shape, while in the bird image, it highlights the distinctive head and wing regions.

This feature-level guidance appears to provide complementary benefits to traditional KD, which focuses on output-level knowledge transfer. 
By explicitly showing the student which input features are most influential for classification, IG augmentation facilitates more efficient learning from limited parameters while maintaining focus on the same discriminative features as the teacher.

\section{Conclusion}\label{sec:conclusions}
Our research introduces an innovative approach to model compression that effectively balances size reduction, performance preservation, and interpretability. 
By combining knowledge distillation with integrated gradients as a data augmentation technique, we demonstrate significant improvements over traditional compression methods.

The experimental results on CIFAR-10 confirm that our IG-enhanced knowledge distillation achieves 92.5\% accuracy with a 4.1x compression factor—representing a substantial 1.02 percentage point improvement over the baseline student model. This compressed model preserves 98.4\% of the performance of the teacher model while delivering a 10.8x reduction in inference time, from 140 ms to 13 ms.

A key innovation in our approach is the pre-computation of IG maps before training, which transforms what would otherwise be a significant computational burden into a one-time preprocessing step. 
This makes our method particularly suitable for practical applications where training efficiency is important.

Our comprehensive ablation studies demonstrate that while both KD and IG independently improve model performance, their combination provides complementary benefits that exceed the sum of their individual contributions. 
This synergy suggests that feature-level guidance through IG effectively addresses limitations in traditional knowledge distillation by explicitly highlighting important regions in the input space.

Looking forward, this research establishes a foundation for developing compression techniques that preserve not only model performance but also interpretability. 
The methodology provides a practical pathway for deploying sophisticated neural networks in resource-constrained environments such as mobile devices and embedded systems, without sacrificing accuracy or explainability.

Future work could explore the application of this approach to more complex datasets and network architectures, as well as the integration of other explainable AI techniques into the compression pipeline. 
The principles established here could also extend beyond classification to other computer vision tasks where model efficiency and interpretability are equally valuable.

\section{Statements and Declarations}

\subsection{Competing Interests}
The authors have no financial or non-financial interests that are directly or indirectly related to the work submitted for publication.

\subsection{Code Availability}
Our models and code are available in the following repository: \url{https://github.com/nordlinglab/ModelCompression-IntegratedGradients}.

\subsection{Authors’ Contribution Statement}
Author contribution using the CRediT taxonomy: 
Conceptualisation: TN and JC; Data curation: DH; Formal analysis: JC and DH; Methodology: JC; Investigation: DH; Software: DH; Verification: DH and JC and TN; Visualisation: DH and JC; Writing - original draft preparation: DH; Writing - review and editing: JC and TN; Funding acquisition: TN; Project administration: TN; Resources: TN; Supervision: JC and TN.

\subsection{Statement of LLM Usage}
This research utilised a large language model (Claude 3.7 Sonnet by Anthropic) to support specific aspects of manuscript preparation. 
Following the completion of our experimental work, model development, and data analysis, we employed Claude 3.7 to help refine the presentation of our methodology and findings. 
The LLM assisted primarily with improving clarity and structure in the Methods and Results sections, helping to articulate complex concepts more precisely. 
All text generated with AI assistance was subsequently reviewed, revised, and validated by the authors to ensure complete accuracy and fidelity to our research. 
No experimental design, figure creation, or technical implementation was performed using AI tools--these core research components were executed exclusively by the authors.


\end{document}